\documentclass[conference]{IEEEtran}
\IEEEoverridecommandlockouts
\usepackage{cite}
\usepackage{amsmath,amssymb,amsfonts}
\usepackage{graphicx}
\usepackage{textcomp}
\usepackage{xcolor}
\usepackage{algorithm}
\usepackage{algpseudocode}
\usepackage{multirow}
\def\BibTeX{{\rm B\kern-.05em{\sc i\kern-.025em b}\kern-.08em
    T\kern-.1667em\lower.7ex\hbox{E}\kern-.125emX}}
\begin{document}

\title{A Comprehensive Study on Quantization Techniques for Large Language Models\\

% {\footnotesize \textsuperscript{*}Note: Sub-titles are not captured in Xplore and should not be used}

}

\author{\IEEEauthorblockN{Jiedong Lang}
\IEEEauthorblockA{\textit{Data Science} \\
\textit{Northeastern University}\\
Boston, USA \\
lang.ji@northeastern.edu}
\and
\IEEEauthorblockN{Zhehao Guo}
\IEEEauthorblockA{\textit{Information Science} \\
\textit{University of Pittsburgh}\\
Pittsburgh, USA \\
zhg26@pitt.edu}
\and
\IEEEauthorblockN{Shuyu Huang}
\IEEEauthorblockA{\textit{Data Science} \\
\textit{Columbia University}\\
New York, USA \\
sh3967@columbia.edu}

}

\maketitle

% introduction: what is quantizaiton, application, first time use(history) -> llm(why use quantization in llm). appliction of quantization in llm, history (related works) 2
% The Two Types of LLM Quantization: PTQ and QAT 0.5
% dive deep to all techniques in quantization LLMs, list all works
%   A: pro, cons -> why b better and cons -> C 3
% conclusion
% 
% 
% 
% 

\begin{abstract}

Large Language Models (LLMs) have been extensively researched and used in both academia and industry since the rise in popularity of the Transformer model, which demonstrates excellent performance in AI. However, the computational demands of LLMs are immense, and the energy resources required to run them are often limited. For instance, popular models like GPT-3, with 175 billion parameters and a storage requirement of 350 GB, present significant challenges for deployment on resource-constrained IoT devices and embedded systems. These systems often lack the computational capacity to handle such large models. Quantization, a technique that reduces the precision of model values to a smaller set of discrete values, offers a promising solution by reducing the size of LLMs and accelerating inference. In this research, we provide a comprehensive analysis of quantization techniques within the machine learning field, with a particular focus on their application to LLMs. We begin by exploring the mathematical theory of quantization, followed by a review of common quantization methods and how they are implemented. Furthermore, we examine several prominent quantization methods applied to LLMs, detailing their algorithms and performance outcomes.
\end{abstract}

% two pages for quantization introduction
\section{Introduction}
Machine Learning has been growing successfully with noticeable rate since 2000, and especially the recent 10 years with a compound growth rate of 30\%. One of the most common branches from machine learning, Deep learning, requires more data than other machine learning branches, which in turn requires more computational power. In general, deep learning models need powerful graphical processing units (GPUs) to handle the large volumes of data and complex calculations involved in training deep neural networks. However, access to the unlimited computational resource (GPUs) for deep learning models is far from ideal due to the high costs. A research field, Quantization in deep learning, aim to reduce the high cost of computations and memory by representing the weights and activation in deep learning models with low precision data types. In the Computer Science, a floating point(float32) consists of 32 bits requires much larger resources than a integer(int8) consists only 8 bits. This type of quantization is straight forward as reducing the number of bits of the data type, from float32 to int8, to consume less computational costs. In addition, mathematics operations including matrix multiplications can be performed with faster speed with lower precision data types. 

% what is quantization and application in Deep learning
The dictionary definition of quantization is the division of a quantity into a discrete number of small parts, often assumed to be integral multiples of a common quantity. The first use of quantization is rounding off and was analyzed by Shappard \cite{gray1998quantization}. In addition, Shannon Entropy quantifies the uncertainty within a dataset, and the process of quantization can influence the calculated entropy by altering the precision of data values\cite{huang2024adversarial}. With the advancement of computer science and machine learning, quantization research fields expands significantly. One of the most common quantization techniques is 8-bit quantization which convert floating point data type to integer data type. While 32-bit single-precision floating-point has been the predominant numerical format for deep learning (DL) applications, alternative formats has emerged recently to enhance the computational performance of these applications\cite{Wu2020IntegerQF}. It is very common to train neural networks using 16-bit floating-point formats, such as fp16 or bfloat16, which are supported by most DL accelerators. After training, neural networks can be deployed for inference using even lower-precision formats, including floating-point, fixed-point, and integer representations. Low-precision formats confer several performance advantages. Firstly, many processors are equipped with higher-throughput mathematical pipelines for low-bit formats, thereby accelerating computation-intensive tasks like convolutions and matrix multiplications. Secondly, reduced word sizes reduce memory bandwidth constraints, resulting in improved performance for bandwidth-limited computations. Third, smaller word sizes decrease memory size requirements, which enhances cache utilization and positively impacts various aspects of memory system performance. Utilizing the int8 quantization, this work achieves the ability to sustain model accuracy within 1\% of the baseline floating-point networks. This is particularly noteworthy for networks that are typically difficult to quantize, such as MobileNets and BERT-large. Moreover, Vector quantization used to compress the deep convolutional networks\cite{gong2014compressing} is a good work supplements this research filed. In general, a CNN that works well object classification contains eight layers and a huge number of parameters, and it is widely known that the parameters are heavily over-parameterized. The goal of this work is to compress these parameters while maintain the high accuracy. This research mainly focus vector quantization methods for the compression of densely connected layers. It involves parameter binarization, scalar quantization through k-means clustering, and structured quantization employing product quantization or residual quantization, all of which lead to significant improvements in performance.

% what is llm, start with all language models
Language model is a branch of machine learning that is designed to understand and generate natural language. Technically, it understands the context of the prompts and generate the missing part with coherent and contextually appropriate language. Language models can be classified into four major types: Statistical Language Models (SLM), Neural Language Models (NLM), Pre-trained Language Models (PLM), and Large Language Models (LLM). Each of these models represents a distinct approach to natural language processing, with varying techniques and capabilities for handling linguistic data\cite{rosenfeld2000two,melis2017state,zhang2022opt,zhao2023survey}. While Statistical Language Models and Neural Language Models have been researched for decades, the Pre-trained Language Models and Large Language Models draw a lot of popularity in both academia and industry recently and the applications are widely used in various fields. For example, recommendation system\cite{li2021extracting,chen2021airec}. Pre-trained Language models propose train largely amount of text data before fine-tune for specific downstream tasks. Based on the Transformer architecture and the self-attention mechanism\cite{vaswani2017attention, fudong:cikm24:mae,fudong:ecml23:storm,fudong:iccv23:mmst_vit,fudong:kdd24:crop_net}, Pre-trained Language Models advances the performance for semantic-purpose language processing. BERT(Bidirectional Encoder Representations from Transformers) serves as an exemplary Pre-trained Language Model, pre-trained on a large corpus of text in an unsupervised manner using masked language modeling and next sentence prediction and fine-tuned for tasks like question-answering, sentiment analysis, and text classification. It has become the dominant paradigm in Pre-trained Language models due to its efficiency, scalability, and superior performance across multiple tasks\cite{koroteev2021bert,,fudong:cikm22:cascade_vae}. Large Language Models represent an extension of research based on Pre-trained Language Models, building upon their foundational Transformer architectures to enhance capabilities in natural language processing. It's found that scaling up PLM model size and data size enhances the capacity to perform more effectively on downstream tasks. For example, GPT-3, a significantly larger PLM with 175 billion parameters, shares a similar architecture and pre-training tasks with standard PLMs. However, despite the primary focus of scaling being on model size, GPT-3 shows remarkable abilities and out performs standard PLM in solving complex tasks. In certain creative writing fields\cite{marco2022systematic}, GPT-3 model highly capable of producing creative content like poetry, song lyrics, or fiction that are coherent to specific styles or themes whereas BERT(340 million parameters) is not designed for creative writing tasks which only capable of completing sentences or predicting missing words. Due to Large Language Model's attribute that requires practical large-scale data processing and distributed parallel training, conducting repetitive studies to explore various training strategies is highly resource-intensive and costly.

% why need of quantization, application of llm quantization, history?
Quantization techniques can help mitigate the high costs associated with training Large Language Models(LLM) by reducing the computational and resource demands. By reducing the number of bits required for each of a model's weights, significantly decreases the overall model size. This reduction leads to LLM that consume less memory, require less storage space, are more energy-efficient, and enable faster inference. These advantages enable LLM to operate on a broader range of devices, including embedded devices and single GPU devices. For instance, supporting AI models on SLAM robotics devices\cite{10503743, xanthidis2021towards, 10684902} or decentralized web3 applications is very challenging, as these systems cannot easily run full-sized models due to their limited capacity for handling high-cost computations\cite{song2024unveiling,gao2023autonomous,gao2024decentralized,song2024advancing}. In such cases, quantization becomes necessary for enabling these integrations, as it reduces model size and computational requirements, making deployment on resource-constrained platforms achievable. While there are various quantization techniques, the two most notable types used in LLM are Post-Traning Quantization(PTQ) and Quantization-Aware Training(QAT). PTQ refers to a technique used to reduce the size and computational demands of a machine learning model after it has been trained and it only affects the inference state. The research work SmoothQuant\cite{pmlr-v202-xiao23c} introduces a PTQ solution aimed at reducing hardware costs and democratizes LLMs. SmoothQuant enables 8-bit weight, 8-bit activation(W8A8) quantization for LLMs and it smooths the activation outliers by offline migrating the quantization difficulty with a mathematically equivalent transformation given that weighs are easy to quantize but activations are not. QuIP\cite{chee2024quip} is another research work that employs PTQ in LLMs. This method is based on the insight that quantization benefits from incoherent weight and Hessian matrices. As a result, QuIP improves several existing quantization algorithms and yields the first LLM quantization methods that produce viable results using only two bits per weight. On the other hand, Quantization-Aware Training (QAT) technique refers to optimize models for efficient inference by simulating the effects of quantization during the training process. In contrast to PTQ techniques, QAT integrates the weight conversion process during the training stage. The Research work Degree-Quant\cite{tailor2021degreequantquantizationawaretraininggraph} efficiently improves inference time of Graph Neural Networks by utilizating the QAT technique. Degree-Quant explores the viability of training quantized GNNs, allowing the use of low precision integer arithmetic during inference. Models trained with Degree-Quant for INT8 quantization perform comparably to FP32 models in most cases, while INT4 models achieve up to a 26\% improvement over baseline models. Moreover, EfficientQAT\cite{chen2024efficientqatefficientquantizationawaretraining} is another research work proposes a more feasible QAT algorithm that satisfies reducing memory consumption during LLM training. EfficientQAT employs a two-step approach: Block-wise training of all parameters (Block-AP) and end-to-end training of quantization parameters (E2E-QP) in which reducing accuracy loss in low-bit scenarios and then trains only the quantization parameters end-to-end. As a result, EfficientQAT outperforms previous quantization methods across a range of models with scales from 7B to 70B parameters at different quantization bit levels. This paper aims to provide a comprehensive review of quantization techniques in the context of LLMs. We begin by detailing the underlying mechanisms of quantization, followed by a comparison of various approaches, with a specific focus on their application at the LLM level.

\section{RANGE MAPPING}
\subsection{AFFINE QUATIZATION}

\subsubsection{THE EQUATION BEHIND THE AFFINE QUATIZATION}The parameter matrices in machine learning models can be exceptionally large. For instance, Chat GPT-3 incorporates approximately 175 billion parameters, represented using 16-bit floating-point precision (float16). Each float16 (FP16) occupies 16 bits, equivalent to 2 bytes. Consequently, the storage requirement for Chat GPT-3 amounts to approximately 350 gigabytes. Quantization is a technique employed to reduce the precision of parameters, thereby optimizing model performance and storage requirements. A widely adopted method for this purpose is the affine quantization scheme, which can be mathematically expressed by the following equation
\[
x_q = (x*S + Z)
\]
\subsubsection{THE PARAMETERS BEHIND THE EQUATION}
Assume the full-precision data range is [$\beta$, $\alpha$]. 

x is the weight without quantization.  

$x_q$ is the quantized weight.

S is the scaling factor, E.g, if converting from FP32 to FP16, would be 2. 

Z is the zero point. It represents in same precision as the quantized value of 0 in full-precision value. z = -round($\beta$*s)-$2^(\beta - 1)$

Quantizing precision within the same data type, such as from FP32 to FP16, is relatively straightforward. However, the process becomes more complex when converting between different data types, such as from FP32 to INT8. The challenge lies in the fact that FP32 represents floating-point numbers, whereas INT8 is limited to integer values. Additionally, FP32 offers a much wider numerical range compared to INT8. To overcome these challenges, a technique known as calibration is employed to determine an appropriate scaling factor, ensuring effective mapping between the two data types. In this case, scaling factor  \[
s = \frac{2^b - 1}{\alpha - \beta}
\]. Once the scaling factor is determined, it can be applied to map unquantized weights to their quantized counterparts effectively.

Through the application of quantization, the size of parameter matrices can be significantly reduced. For example, in the case of Chat GPT-3, if the precision is reduced from FP16 to FP4, each parameter’s size would decrease by a factor of four, according to the previously discussed equation. As a result, the total storage requirement for the 175 billion parameters in the Chat GPT-3 model would be reduced to approximately 90 gigabytes, representing a substantial reduction compared to the full-precision model.
\hfill\\
\subsection{SCALE QUATIZATION}
\subsubsection{THE EQUATION BEHIND THE SCALE QUATIZATION}
Let us assume the range of full-precision value is [-$\alpha$, $\alpha$] and it will be quantized to a b-bit integer value x.
\[
s = \frac{2^{b-1} - 1}{\alpha}
\]
To quantize a floating data:
\[
x_q = (x*S)
\]
\subsubsection{DETAILS OF SCALE QUANTIZATION}
Different from affine quantization, scale quantization only performs quantization with a scale transformation. This is usually used for two data types have equal proportion and same zero points. \cite{hwu2020integer}. For simplicity and symmetry, the range of int8 would be [-127, 127], instead of [-127,128], so the range would symmetric. But for lower bit representation, ignoring one number for symmetry might shorten the mapping value range. 
\hfill\\

\section{QUANTIZATION GRANULARITY}
The granularity of quantization can be decided based on the requirement of training the model.  
\hfill\\

\textbf{PER-LAYER QUANTIZATION}
n a neural network, all filters within the same layer share the same quantization parameters, with the quantization range determined collectively based on the values across all filters in that layer. This level of quantization granularity simplifies the quantization process but often results in reduced model performance. The primary reason is that applying a uniform quantization strategy to all filters may introduce larger quantization errors for individual filters, compromising the model's overall accuracy.

\textbf{PER-CHANNEL QUANTIZATION}
To enhance performance following quantization, a higher granularity of quantization is employed. Instead of being applied at the layer level, quantization is performed at the filter level within each layer of the neural network. Each filter is assigned customized quantization parameters, tailored to the specific range of values within that filter. Although this approach increases the complexity of the quantization process, it generally leads to improved model performance compared to methods with lower granularity, as it reduces quantization errors for individual filters.

\section{QUANTIZATION CALIBRATION}

The maximum and minimum values from the full-precision data range are utilized to scale the data to a lower precision. For example, when scaling from INT16 to INT8, the INT16 value range of $[-32767, 32767]$ is evenly mapped onto the INT8 value range of $[-125, 125]$. However, various methods exist to determine the optimal value range prior to quantization, aiming to enhance the overall performance of the quantized model.

\textbf{GLOBAL CALIBRATION}
This approach represents the simplest method of calibration. It selects the maximum and minimum values from the unquantized data without distinction and converts them to a lower precision. While this method simplifies the quantization process, it can compromise accuracy, as not all values within the unquantized data range are necessarily relevant or required for optimal performance.

\textbf{MAX CALIBRATION}
This calibration method selects the maximum value from the unquantized data, rather than relying on the global value range of the unquantized data type. This approach enables a more precise alignment between the quantization process and the parameters, thereby mitigating the loss typically associated with quantization.

\textbf{KL DIVERGENCE CALIBRATION}
Kullback-Leibler (KL) divergence measures the difference between two probability distributions. In the context of quantization, it is employed to compare the distribution of quantized data with that of the original full-precision data. Various scaling factors are evaluated to generate different distributions for the full-precision data, and KL divergence is used to identify the scaling factor that minimizes information loss. This approach enables more effective quantization compared to max-value calibration, as it preserves data integrity more accurately.

\textbf{PERCENTILE CALIBRATION}
Percentile calibration is based on the distribution of the data rather than the full range of the original precision data type. For instance, in the case of normally distributed data, the minimum and maximum values are sparsely located at the distribution’s tails. Percentile calibration focuses on a specified percentile, excluding values that fall below or above certain thresholds. This approach effectively narrows the range of full-precision values, thereby improving the performance of quantization. Optimal percentile values for calibration are typically 99.99\% or 99.999\%, as lower percentiles are considered too aggressive, excluding more high-magnitude values and negatively impacting quantization performance \cite{wu2020integer}.
\hfill\\
\section{QUANTIZATION TECHNIQUES}
There are several methods available for implementing model quantization. This section presents a range of quantization techniques aimed at achieving faster computation, minimizing accuracy loss, and reducing the model's parameter size.

\textbf{post-training Quantization}
The primary concept of post-training quantization is to apply quantization after the model has been fully trained. Two common approaches are used to achieve this:
Dynamic Quantization: In this method, quantization occurs at runtime after each activation. However, this approach introduces additional computational overhead, potentially slowing down performance due to the extra processing required for each activation.
Static Quantization: Here, the quantization parameters are pre-computed during the quantization process, prior to runtime. This method ensures the quantization of weights while minimizing runtime overhead, as the need for on-the-fly calculations is eliminated.

\textbf{Quantization-Aware Training}
The model accounts for the errors introduced by quantization by incorporating quantization operators at each activation during the training phase. These operators enable the model to recognize and adapt to quantization errors throughout the backpropagation process. As a result, this approach typically leads to reduced performance degradation while also facilitating faster computation \cite{gholami2021surveyquantizationmethodsefficient}.

\textbf{Weight Quantization}
Instead of quantizing all parameters in the model, this technique selectively quantizes only a subset. Specifically, it targets the weight matrices for quantization while preserving the activation values in their original precision. By employing this approach, the data storage requirements can be significantly reduced.

\textbf{Activation-Aware Weight Quantization}
While traditional weight quantization applies to all parameters within the weight matrices, alternative approaches can further reduce the number of parameters to be quantized, thereby enhancing quantization speed. Activation-Aware Weight Quantization (AWQ) \cite{lin2024awq} introduces a selective approach, recognizing that not all weights are equally important and only a subset requires quantization. This technique identifies critical weights based on activation magnitudes, retaining these essential weights in full precision while quantizing the non-critical ones. By focusing on the most influential parameters, this method minimizes accuracy loss while maintaining lower computational costs.

\textbf{Attention-Aware Weight Quantization}
This quantization technique utilizes the Hessian trace as a measure to determine the importance of weight matrices \cite{guan2024aptq}. It leverages the attention mechanism, commonly employed in Large Language Models (LLMs). Weights identified as more significant based on attention scores are assigned higher-bit precision, while less important weights are quantized using lower-bit representations. This approach adopts mixed-precision quantization, resulting in improved performance compared to previous quantization techniques. Notably, it achieves superior efficiency in LLMs without compromising model accuracy.

% dive deep to all techniques in quantization LLMs, list all works
%   A: pro, cons -> why b better and cons -> C 3
% 3 PQT
%   GPTQ https://arxiv.org/pdf/2210.17323 

% 1 QAT
% https://arxiv.org/pdf/2305.17888 LLM-QAT: Data-Free Quantization Aware Training  - read this and write a detailed summary

\section{Analysis of Quantization Approaches in LLMs}
After exploring various quantization techniques and the theory behinds, we are going to focus more on the two most common used quantization methods in LLMs, Post-Training Quantization (PTQ) and Quantization-Aware Training (QAT). While both techniques enable models to operate in lower-precision formats, they differ in their trade-offs between ease of implementation, computational cost, and model accuracy. PTQ is widely favored for its simplicity and fast deployment, as it can be applied to a pre-trained model without requiring additional training. However, for large and complex models like LLMs, PTQ can result in considerable accuracy loss, since the model is not optimized for quantization. On the other hand, QAT involves simulating quantization during the training process, allowing the model to adjust to lower precision. Although this method results in better accuracy retention, it introduces additional computational overhead and longer training times. In this section, we dive deep into two noticeable works that one utilizing PTQ while another one employing QAT. We will examine their respective improvements and limitations in detail.

\begin{figure}[htp]  
\centering
\includegraphics[width=0.9\columnwidth]{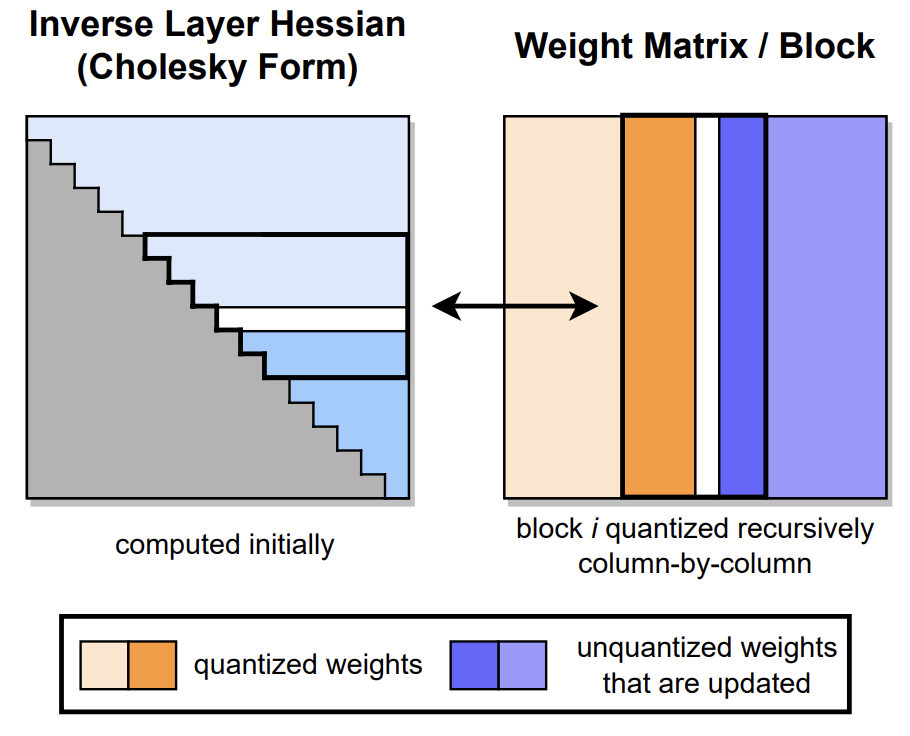}
\caption{GPTQ quantizes the weights in blocks, with each block consisting of 128 columns, similar to a sliding window approach. This block-wise quantization allows the model to process smaller segments of weights at a time, improving computational efficiency while maintaining accuracy during the quantization process.}
\label{fig:gptq}
\end{figure}

\hfill\\
\textbf{GPTQ.} is a quantization technique utilizing PTQ designed to reduce the size of models without great loss of accuracy on LLMs\cite{frantar2022gptq}. Previously, a research work, an Extreme Compression for Pre-trained Transformers Made Simple and Efficient, applies compression at the scale of GPT-175B \cite{wu2022xtc}. The compression works well for low compression targets, e.g., 8-bit weights, but the work fails to preserve accuracy at higher rate and leaves question that if one-shot PTQ with higher accuracy a feasible approach. GPTQ addresses the challenge of efficient execution for models with hundreds of billions of parameters by introducing a novel PTQ method. This approach enables models to be executed within a few hours while compressing them to 3 or 4 bits per parameter without significant accuracy loss. Additionally, GPTQ supports extreme quantization, successfully reducing models to as few as 2 bits. In practice, this allows the compressed OPT-175B model to run on a single NVIDIA A100 GPU or two NVIDIA A6000 GPUs. By implementing custom GPU kernels that leverage compression for faster memory loading, GPTQ achieves notable speed improvements—around 3.25x on A100 GPUs and 4.5x on A6000 GPUs. However, its speed gains are limited during multiplication operations due to the restricted hardware capabilities for mixed-precision operands (e.g., FP16 x INT4) on mainstream architectures. Before delving into the algorithm, we first review the background of the research. This study implements Layer-wise Quantization, which performs quantization on a layer-by-layer basis and addresses a corresponding reconstruction problem within the PTQ technique. In simple terms, the objective is to find a matrix of quantized weights that minimizes the error in relation to the full-precision output of each layer. It can be formulated as a mathematical function as 
\begin{equation}
\arg\min_{\widehat{W}} \| WX - \widehat{W}X \|_2^2
\label{eq:argmin}
\end{equation}
where $W$ denotes weights and $X$ denotes the layer input. In addition, this research build on a work called Optimal Brain Quantization (OBQ) which aims to find the minimum error on each layer as we described earlier. As in Equation~\eqref{eq:argmin}, OBQ solves by quantizing each row \textbf{w} of $W$ independently and always updating the rest not quantized weights, in which the Hessian matrix is $ \mathbf{H}_F = 2 \mathbf{X}_F \mathbf{X}_F^\top $, $F$ denotes the remaining full precision weights. 

\begin{equation}
\begin{split}
w_q = \arg\min_{w_q} \frac{(\text{quant}(w_q) - w_q)^2}{\left[\mathbf{H}_F^{-1}\right]_{qq}}, \\
\delta_F = - \frac{w_q - \text{quant}(w_q)}{\left[\mathbf{H}_F^{-1}\right]_{qq}} 
           \cdot \left( \mathbf{H}_F^{-1} \right)_{:,q}
\end{split}
\label{eq:eq2}
\end{equation}
As defined in Equation~\eqref{eq:eq2}, $w_q$ denotes the optimal weight to be quantized next, and the corresponding weights to be updated in $F$ is denoted as $\delta_F$. OBQ repeatedly follows these equations during the quantization of weights, effectively solving them for medium-sized models. However, as model sizes grow larger (reaching billions of parameters), the computational cost increases significantly. To address this, GPTQ introduces modifications to the quantization procedure to improve computational efficiency. Instead of independently quantizing the weights of each row based on their corresponding errors, GPTQ quantizes the weights of all rows in the same order. This approach yields a final squared error comparable to the greedy optimal solution (OBQ). As a result, this method reduces computations on the set of full-precision weights $F$ and $\mathbf{H}_F^{-1}$, since they remain the same for all rows. As illustrated in Figure\ref{fig:gptq}, GPTQ quantizes the weights column by column and this provides more than three orders magnitude computational speedup. 

\textbf{GPTQ Full Algorithm} presented in pseudo-code as seen in Algorithm\ref{alg:pseudoGPTQ} \cite{frantar2022gptq}
\begin{algorithm}
\caption{Quantize $W$ given inverse Hessian $\mathbf{H}^{-1} = (2\mathbf{X}\mathbf{X}^\top + \lambda \mathbf{I})^{-1}$ and blocksize $B$.}
\begin{algorithmic}[1]
\State \textbf{Q} $\gets 0_{d_{\text{row}} \times d_{\text{col}}}$ \Comment{quantized output}
\State \textbf{E} $\gets 0_{d_{\text{row}} \times B}$ \Comment{block quantization errors}
\State $\mathbf{H}^{-1} \gets \text{Cholesky}(\mathbf{H}^{-1})^\top$ \Comment{Hessian inverse information}
\For{$i = 0, B, 2B, \dots$}
    \For{$j = i, \dots, i + B - 1$}
        \State $\textbf{Q}_{:, j} \gets \text{quant}(\textbf{W}_{:, j})$ \Comment{quantize column}
        \State $\textbf{E}_{:, j-i} \gets (\textbf{W}_{:, j} - \textbf{Q}_{:, j}) / \left[ \mathbf{H}^{-1} \right]_{jj}$ \Comment{quantization error}
        \State $\textbf{W}_{:, j:(i+B)} \gets \textbf{W}_{:, j:(i+B)} - \textbf{E}_{:, j-i} \cdot \mathbf{H}^{-1}_{j, j:(i+B)}$ \Comment{update weights in block}
    \EndFor
    \State $\textbf{W}_{:, (i+B):} \gets \textbf{W}_{:, (i+B):} - \textbf{E} \cdot \mathbf{H}^{-1}_{i:(i+B), (i+B):}$ \Comment{update all remaining weights}
\EndFor
\end{algorithmic}
\label{alg:pseudoGPTQ}
\end{algorithm}

In the practice, the research validates GPTQ's performance against existing quantization methods, focusing on both smaller models, ResNet and BERT, and large-scale models, BLOOM and OPT. For quantization in small models, GPTQ performs comparably to other methods at 4-bit precision but is slightly less accurate than the most precise techniques at 3-bit precision. Despite this, GPTQ significantly outperforms AdaQuant—the fastest PTQ method among those compared—in terms of speed. Overall, GPTQ is highly competitive with state-of-the-art PTQ methods for smaller models, reducing processing time from hours to less than a minute. GPTQ successfully compresses the entire OPT and BLOOM models to 3-bit and 4-bit precision. The performance of these quantized models is evaluated across several language tasks, including WikiText2. As shown in Table \ref{tab:OPTCompare} and Table \ref{tab:BLOOMCompare}, GPTQ significantly outperforms RTN, with RTN showing a notable drop in performance at 3-bit precision for the OPT model. Overall, GPTQ successfully introduces an effective approach for quantizing Large Language Models (LLMs), enabling the accurate quantization of the largest public models to 3-bit and 4-bit precision with minimal accuracy loss. While the research demonstrates superior performance in Post-Training Quantization (PTQ), its focus is limited to generative tasks, with no exploration of activation quantization.
% Table 4: BLOOM perplexity results for WikiText2
\begin{table}[h!]
\centering
\begin{tabular}{|c|c|c|c|c|c|c|c|c|c|}
\hline
\multirow{2}{*}{\textbf{BLOOM}} & \multirow{2}{*}{\textbf{Bits}} & \multicolumn{6}{c|}{\textbf{Models}}
\\ \cline{3-8} 
 &  & \textbf{560M} & \textbf{1.1B} & \textbf{1.7B} & \textbf{3B} & \textbf{7.1B} & \textbf{176B} \\ \hline
\textbf{full} & \textbf{16} & 22.42 & 17.69 & 15.39 & 13.48 & 11.37 & 8.11 \\ \hline
\multirow{2}{*}{\textbf{RTN}} & \textbf{4} & 25.90 & 22.00 & 16.97 & 14.76 & 12.10 & 8.37 \\ \cline{2-8} 
 & \textbf{3} & 57.08 & 50.19 & 63.59 & 39.36 & 17.38 & 571 \\ \hline
\multirow{2}{*}{\textbf{GPTQ}} & \textbf{4} & \textbf{24.03} & \textbf{19.05} & \textbf{16.48} & \textbf{14.20} & \textbf{11.73} & \textbf{8.21} \\ \cline{2-8} 
 & \textbf{3} & \textbf{32.31} & \textbf{25.08} & \textbf{21.11} & \textbf{17.40} & \textbf{13.47} & \textbf{8.64} \\ \hline
\end{tabular}
\vspace{10pt}
\caption{BLOOM perplexity results for WikiText2.}
\label{tab:BLOOMCompare}
\end{table}

\textbf{LLM-QAT}, Key-Value Data-Free Quantization Aware Training, is a specialized quantization technique designed to reduce quantization errors in the Key-Value (KV) caches of Large Language Models (LLMs). KV caches store critical intermediate outputs from attention layers, ensuring the model does not need to recompute information for each token. Unlike other quantization methods that primarily focus on activations or weights, KV-QAT targets these caches, which play a crucial role in every token's processing. While post-training quantization is commonly employed in LLMs, Quantization-Aware Training (QAT) is underutilized due to the resource-intensive nature of LLM training and the limited availability of suitable training data. A notable limitation of post-training quantization is the significant degradation in performance when precision drops below 8 bits \cite{liu2023llmqatdatafreequantizationaware}. To address these challenges, KV-QAT emphasizes the quantization of KV caches and employs knowledge distillation to overcome the lack of training data, particularly when the data is sensitive, restricted, or unavailable.
\begin{table*}[h!]
\centering
    
\begin{tabular}{|l|c|c|c|c|c|c|c|c|c|}  % Add | for vertical lines
    \hline  % Top horizontal line
    \textbf{Method} & \textbf{\#Bits} & \textbf{Size (GB)} & \textbf{BoolQ} & \textbf{PIQA} & \textbf{SIQA} & \textbf{HellaSwag} & \textbf{WinoGrande} & \textbf{ARC-e} & \textbf{Avg.} \\
    \hline  % Horizontal line
    LLaMA-7B & 16-16-16 & 12.6 & 76.8 & 79.3 & 48.6 & 76.1 & 70.0 & 73.0 & 66.2 \\
    \hline
    RTN & 4-8-4 & 3.5 & 51.9 & 56.3 & 40.5 & 35.7 & 49.9 & 39.3 & 41.2 \\
    SmoothQuant & 4-8-4 & 3.5 & 54.7 & 55.4 & 41.1 & 38.9 & 51.5 & 43.9 & 43.2 \\
    LLM-QAT & 4-8-4 & 3.5 & 69.5 & 75.4 & 46.6 & 69.2 & 64.6 & 66.0 & \textbf{60.7} \\
    \hline
    RTN & 4-8-8 & 3.5 & 67.8 & 76.6 & 47.2 & 71.4 & 67.2 & 67.4 & 61.8 \\
    SmoothQuant & 4-8-8 & 3.5 & 71.0 & 76.0 & 45.4 & 67.8 & 66.0 & 67.4 & 60.4 \\
    LLM-QAT & 4-8-8 & 3.5 & 74.6 & 77.5 & 48.3 & 73.5 & 67.7 & 70.2 & \textbf{64.2} \\
    \hline
    RTN & 4-6-16 & 3.5 & 62.4 & 74.5 & 46.8 & 69.7 & 64.5 & 64.6 & 58.9 \\
    SmoothQuant & 4-6-16 & 3.5 & 68.8 & 73.9 & 44.5 & 65.7 & 65.3 & 66.0 & 59.5 \\
    LLM-QAT & 4-6-16 & 3.5 & 72.9 & 76.8 & 47.9 & 72.4 & 68.3 & 68.8 & \textbf{63.1} \\
    \hline
    RTN & 4-8-16 & 3.5 & 67.6 & 77.4 & 47.1 & 71.6 & 67.1 & 67.1 & 61.9 \\
    SmoothQuant & 4-8-16 & 3.5 & 70.2 & 76.4 & 44.8 & 68.1 & 66.0 & 67.3 & 60.6 \\
    LLM-QAT & 4-8-16 & 3.5 & 74.8 & 77.8 & 48.6 & 73.6 & 69.0 & 69.7 & \textbf{64.4} \\
    \hline
    RTN & 4-16-16 & 3.5 & 71.2 & 77.3 & 47.6 & 72.7 & 66.9 & 68.8 & 61.9 \\
    GPTQ & 4-16-16 & 3.5 & 67.7 & 76.0 & 43.0 & 69.4 & 66.7 & 66.9 & 60.9 \\
    LLM-QAT & 4-16-16 & 3.5 & 75.5 & 78.3 & 48.4 & 74.7 & 69.0 & 70.0 & \textbf{64.4} \\
    \hline
    RTN & 8-8-4 & 6.5 & 54.7 & 59.4 & 43.1 & 45.6 & 51.2 & 29.6 & 57.4 \\
    SmoothQuant & 8-8-4 & 6.5 & 60.7 & 67.5 & 44.9 & 58.3 & 58.6 & 36.9 & 58.6 \\
    LLM-QAT & 8-8-4 & 6.5 & 71.1 & 75.6 & 47.3 & 71.8 & 66.3 & 43.6 & \textbf{61.6} \\
    \hline
    RTN & 8-8-8 & 6.5 & 76.4 & 79.5 & 48.7 & 75.5 & 69.7 & 72.3 & 65.9 \\
    SmoothQuant & 8-8-8 & 6.5 & 76.1 & 79.6 & 48.7 & 75.7 & 70.1 & 73.7 & \textbf{66.3} \\
    LLM-QAT & 8-8-8 & 6.5 & 76.0 & 79.6 & 48.5 & 75.7 & 69.4 & 73.1 & 66.0 \\
    \hline
    RTN & 8-8-16 & 6.5 & 76.4 & 79.1 & 48.3 & 75.5 & 69.5 & 72.8 & 65.9 \\
    SmoothQuant & 8-8-16 & 6.5 & 76.2 & 79.5 & 48.6 & 76.1 & 70.5 & 73.2 & \textbf{66.1} \\
    LLM-QAT & 8-8-16 & 6.5 & 76.3 & 79.4 & 48.7 & 75.6 & 69.7 & 72.3 & 65.7 \\
    \hline
\end{tabular}
\vspace{10pt}
    \caption{performance comparison of different quantization methods}
    \label{table:performance}
\end{table*}
To mitigate the data scarcity issue for QAT, KV-QAT uses self-generated data from the model itself. Data-free methods do not rely on the original training datasets but instead generate synthetic data to fine-tune the model, helping it adapt to quantization-induced errors. The data used for this purpose is derived from the next token predicted by the model. However, if the data generation strategy selects only the token with the highest probability, it can result in limited training data with low diversity, as generated sentences tend to be repetitive.
To overcome this issue, a more effective strategy involves randomly selecting the next token based on the probability distribution predicted by the model. This distribution is calculated using the softmax function, which transforms the model’s predictions into a list of possible tokens, with higher probabilities indicating a greater likelihood of selection. This random sampling approach increases the variety and size of the training data, improving the model's robustness to quantization errors.
% Table 3: OPT perplexity results on WikiText2
\begin{table*}[h!]
\centering
\begin{tabular}{|c|c|c|c|c|c|c|c|c|c|c|}
\hline
\multirow{2}{*}{\textbf{OPT}} & \multirow{2}{*}{\textbf{Bits}} & \multicolumn{9}{c|}{\textbf{Models}} \\ \cline{3-11} 
 &  & \textbf{125M} & \textbf{350M} & \textbf{1.3B} & \textbf{2.7B} & \textbf{6.7B} & \textbf{13B} & \textbf{30B} & \textbf{66B} & \textbf{175B} \\ \hline
\textbf{full} & \textbf{16} & 27.65 & 22.00 & 14.63 & 12.47 & 10.86 & 10.13 & 9.56 & 9.34 & 8.34 \\ \hline
\multirow{2}{*}{\textbf{RTN}} & \textbf{4} & 37.28 & 25.94 & 48.17 & 16.92 & 12.10 & 11.32 & 10.98 & 110 & 10.54 \\ \cline{2-11} 
 & \textbf{3} & 1.3e3 & 64.57 & 1.3e4 & 1.6e4 & 5.8e3 & 3.4e3 & 1.6e3 & 6.1e3 & 7.3e3 \\ \hline
\multirow{2}{*}{\textbf{GPTQ}} & \textbf{4} & \textbf{31.12} & \textbf{24.24} & \textbf{15.47} & \textbf{12.87} & \textbf{11.39} & \textbf{10.31} & \textbf{9.63} & \textbf{9.55} & \textbf{8.37} \\ \cline{2-11} 
 & \textbf{3} & \textbf{53.85} & \textbf{33.79} & \textbf{20.97} & \textbf{16.88} & \textbf{14.86} & \textbf{11.61} & \textbf{10.27} & \textbf{14.16} & \textbf{8.68} \\ \hline
\end{tabular}
\vspace{10pt}
\caption{OPT perplexity results on WikiText2.}
\label{tab:OPTCompare}
\end{table*}
For quantizing the KV caches in LLMs, KV-QAT applies MinMax quantization to both weights and activations, ensuring efficient compression without significant performance degradation \cite{liu2023llmqatdatafreequantizationaware}.
\begin{equation}
\begin{split}
\mathbf{X}_Q^i = \alpha \left[ \frac{\mathbf{X}_R^i}{\alpha} \right], \quad
\alpha = \frac{\max\left( |\mathbf{X}_R| \right)}{2^{N-1} - 1}
\end{split}
\label{eq:eq3}
\end{equation}
Here, $\mathbf{X}_Q$ denotes the quantized weights and activations, and $\mathbf{X}_R$ denotes the full-precision weights and activation.Additionally, per-token activation quantization and per-channel weight quantization are employed, as illustrated in Figure 2. Per-token quantization refers to the process of quantizing the key-value pairs in the cache on a token-by-token basis, applying quantization individually to each token’s data as it is generated. Furthermore, the quantization process is integrated with gradient computation, ensuring that the model learns to effectively handle quantized key-value pairs during training. During the QAT process, quantization is applied to the activation tensors associated with both the keys and values, enabling the model to adapt to the quantized representations throughout training. 
Table 3 \cite{liu2023llmqatdatafreequantizationaware} presents a comparison of various quantization techniques under different precision settings for weights, activations, and Key-Value (KV) pairs in the cache. The results demonstrate that, among Round-To-Nearest (RTN), SmoothQuant, and LLM-QAT, the accuracy of LLM-QAT surpasses the other two techniques when the quantization precision drops below 8 bits. Furthermore, even with precision levels above 8 bits, LLM-QAT maintains accuracy comparable to the other methods. These findings indicate that, unlike prior approaches that do not account for KV cache quantization and experience significant accuracy degradation at lower precision levels, LLM-QAT achieves better performance with minimal accuracy loss, particularly in the sub-8-bit precision range.
The introduction of LLM-QAT enhances memory efficiency by enabling extremely large language models to better manage memory through the quantization of key-value (KV) caches with minimal accuracy loss. By compressing the KV cache overhead, these models can accommodate longer sequences and larger batch sizes during training. Additionally, the compressed KV cache improves throughput, facilitating faster inference with increased batch sizes and extended context lengths. Furthermore, LLM-QAT addresses the challenge of limited access to training data. It leverages data generated by pre-trained models, eliminating the need for direct access to the original datasets, thereby ensuring effective fine-tuning without compromising performance. 

\section{Conclusion}
After thoroughly examining various popular quantization techniques, it is evident that their performance varies depending on the precision of the quantization. For example, GPTQ achieves optimal performance at 4-bit precision but experiences a decline at 3-bit precision. In addition, LLM-QAT demonstrates better accuracy with a configuration of 4-bit weights, 4-bit KV caches, and 8-bit activations, compared to a uniform 4-bit setting across all precision \cite{liu2023llmqatdatafreequantizationaware}.
To maximize the performance of a given quantization technique, developers must carefully select appropriate precision settings. Future research on quantization techniques could further explore the impact of precision configuration, potentially leading to more refined and efficient quantization strategies.
\begin{figure}
    \centering
    \includegraphics[scale=0.25]{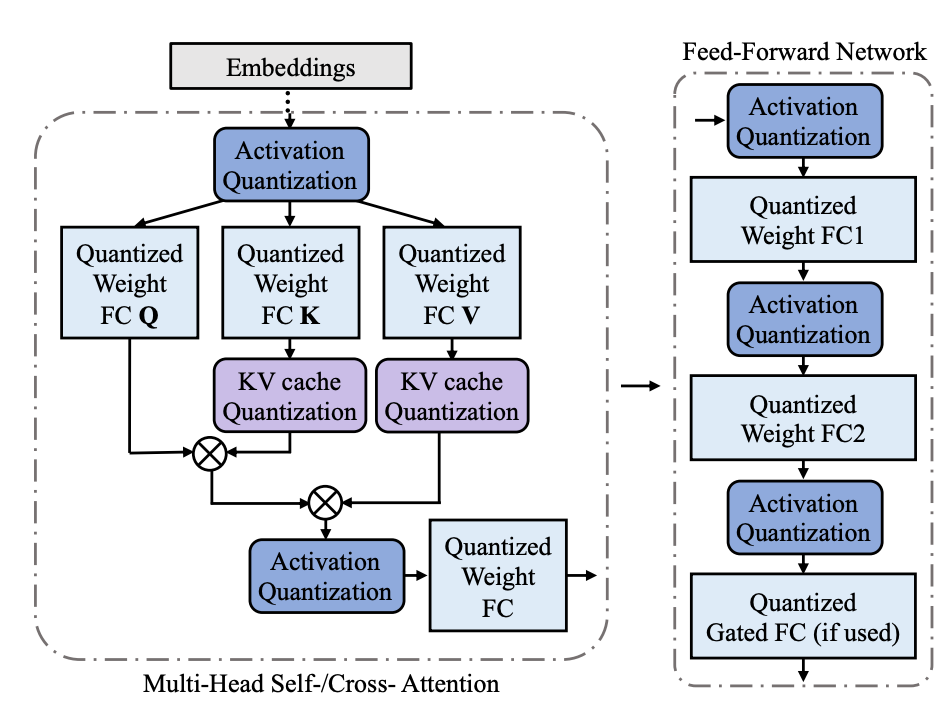}
    \caption{}
    \label{figure 2}
\end{figure}

\bibliographystyle{IEEEtran}
\bibliography{references}

\end{document}